\documentclass[10pt,twocolumn,letterpaper]{article}

\usepackage[pagenumbers]{cvpr} % To force page numbers, e.g. for an arXiv version

\usepackage{graphicx}
\usepackage{multirow}
\usepackage{bbding}
\usepackage{amssymb}

\usepackage[dvipsnames]{xcolor}

\usepackage[pagebackref,breaklinks,colorlinks]{hyperref}

\usepackage[capitalize]{cleveref}
\crefname{section}{Sec.}{Secs.}
\Crefname{section}{Section}{Sections}
\Crefname{table}{Table}{Tables}
\crefname{table}{Tab.}{Tabs.}

\def\wacvPaperID{1474} % *** Enter the WACV Paper ID here
\def\confName{WACV}
\def\confYear{2024}

\title{mmFUSION: Multimodal Fusion for 3D Objects Detection}

\author{
    Javed Ahmad$^{1,2}$ \quad Alessio Del Bue $^1$ \\
    $^1$Pattern Analysis \& Computer Vision (PAVIS), Istituto Italiano di Tecnologia (IIT)\\ \quad $^2$ Universita degli Studi di Genova \\
    {\tt\small javed.ahmad@iit.it, alessio.delbue@iit.it}
}
\begin{document}
\maketitle
\begin{abstract}

Multi-sensor fusion is essential for accurate 3D object detection in self-driving systems. Camera and LiDAR are the most commonly used sensors, and usually, their fusion happens at the early or late stages of 3D detectors with the help of regions of interest (RoIs). On the other hand, fusion at the intermediate level is more adaptive because it does not need RoIs from modalities but is complex as the features of both modalities are presented from different points of view.
In this paper, we propose a new intermediate-level multi-modal fusion (mmFUSION) approach to overcome these challenges. First, the mmFUSION uses separate encoders for each modality to compute features at a desired lower space volume. Second, these features are fused through cross-modality and multi-modality attention mechanisms proposed in mmFUSION. The mmFUSION framework preserves multi-modal information and learns to complement modalities' deficiencies through attention weights. The strong multi-modal features from the mmFUSION framework are fed to a simple 3D detection head for 3D predictions. We evaluate mmFUSION on the KITTI and NuScenes dataset where it performs better than available early, intermediate, late, and even two-stage based fusion schemes.
The code with the mmdetection3D\cite{contributors2020mmdetection3d} project plugin will be publicly available soon.

\end{abstract}    
\section{Introduction}
\label{sec:intro}

\begin{figure}[ht!]
  \centering
\includegraphics[width=14cm,height=8.5cm,keepaspectratio]{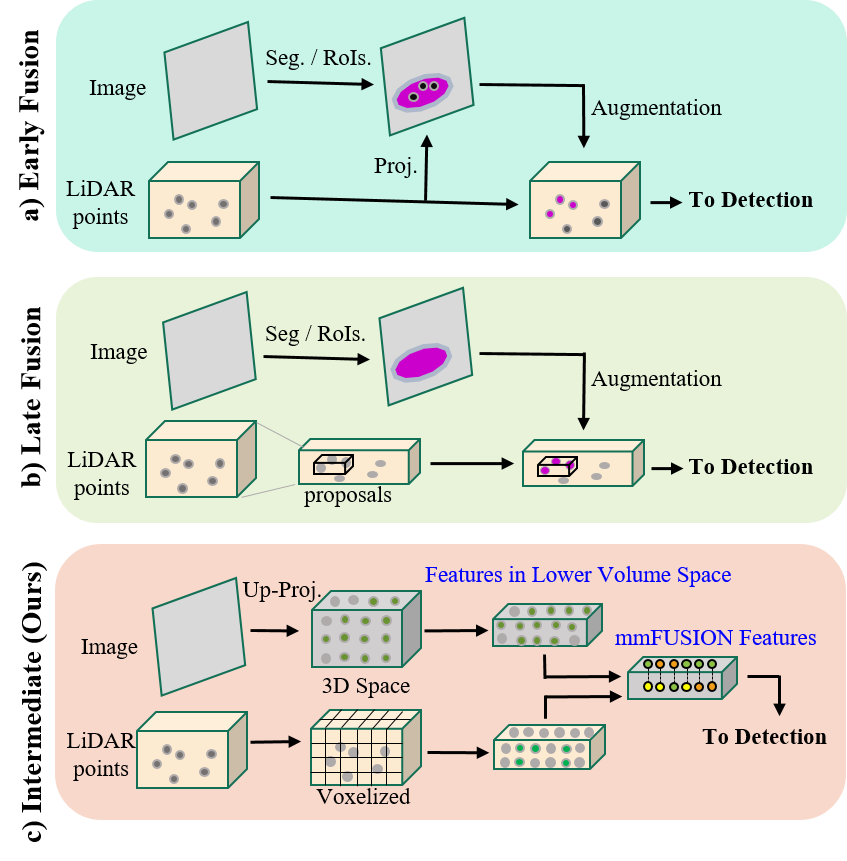}
\caption{Different fusion schemes in 3D detection pipeline. \textbf{(a) Early Fusion:} Low-level multi-modal features are associated at an early stage with the help of RoIs before predicting final 3D proposals. \textbf{(b) Late Fusion:} Proposals from a particular modality are predicted first, and then high-level features or proposals from both modalities interact to get final detection. \textbf{(c) Intermediate stage (Our Case)}: Instead of using RoIs or proposals, we encode each modality features separately at a defined lower space volume, fuse these feature through attention weighing, and compute joint strong multi-modal features. }
  \label{fig:intro}
\end{figure} 
% A gentle 3d detection intro, touch, and effect of fusion strategies on detection
%
% a gentle introduction
3D object detection is a fundamental task in autonomous driving systems and robotics while being a fundamental skill to enable further 3D scene understanding applications. These systems are equipped with different sensors such as cameras, LiDARs, and radars. Every sensor has its merits and drawbacks, for instance, RGB cameras provide rich semantic information while lacking scene depth information; whereas depth sensors, such as LiDARs, are accurate in terms of 3D localization but they lack in providing a complete geometry of the scene \cite{lang2019pointpillars, zhou2018voxelnet, yang2019std, shi2019pointrcnn} given their sparse output. Therefore, it is of great importance to extract complementary information from these diverse sensors and fuse them for better accuracy and reliable perception.

% introducing early, late and intermediate fusion schemes
The process of associating and fusing multi-modality features in the 3D detection pipeline can be categorized as early, late, and intermediate-stage fusion. Early fusion is a method of augmenting LiDAR points with image semantic labels and features based on RoIs (see Figure \ref{fig:intro}a). Even the early fusion is a well-known strategy on large-scale detection benchmarks, but its performance is restricted because of a very limited number of camera features matched to sparse LiDAR points \cite{vora2020pointpainting,liang2018deep,huang2020epnet,yin2021multimodal,li2022voxel}. The late fusion schemes first predict coarse-level proposals from at least one modality and then interact with other modality features for final 3D prediction (see  Figure \ref{fig:intro}b). This fusion scheme is predominant because of its simplicity, but the overall performance is limited if the initial proposals are based on the sensor with deficiencies \cite{chen2017multi,ku2018joint,yoo20203d}. The intermediate-level fusion works on high-level features computed from each modality and can not depend on RoIs as the other fusion schemes. This fusion scheme is more adaptable for multi-sensors \cite{liang2018deep, yoo20203d} but faces several challenges in transforming diverse multi-modality features into a common volume space.
Usually, the transformation process misaligns the modalities' features which degrades the performance of 3D detectors. Hence we propose a new multi-modal fusion method that transforms each modality feature into a common feature space retaining meaningful information. Subsequently, complement these features with an attention mechanism that resolves the alignment problem during the fusion. (see in Figure \ref{fig:intro}c).

\begin{figure*}[t]
  \centering
  \includegraphics[width=17.5cm,height=9.2cm,keepaspectratio]{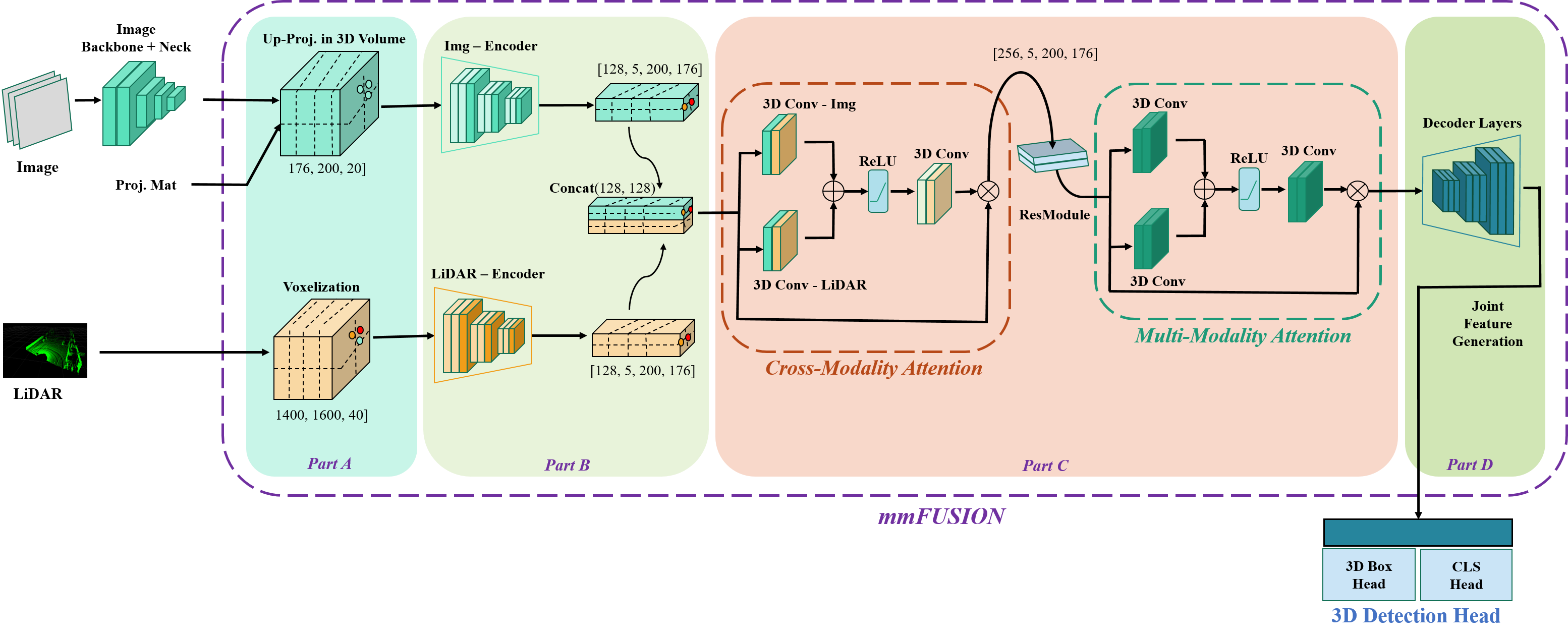}
  \caption{The framework for 3D object detection with mmFUSION. The \textbf{Image} and \textbf{LiDAR }point clouds features are extracted using their respective \textbf{backbones} and these features are transformed to respective defined 3D volumes as shown in part \textcolor{violet}{\textit{Part A}}. In \textcolor{violet}{\textit{Part B}}: the \textbf{Img-Encoder} and \textbf{LiDAR-Encoder} transform these features to a lower size volume preserving all significant information. Both representations are concatenated at the feature dimension. In \textcolor{violet}{\textit{Part C}}: these features are processed through \textbf{cross-modality attention}, where attention learned weights are multiplied by incoming features of both modalities. Further, after passing through a \textbf{residual block}, these multi-modal features are weighted by \textbf{multi-modal attention}. Afterward, in part \textcolor{violet}{\textit{Part D}}, the decoder layers generate desired the final joint features. The following block is the 3D detection Head to regress boxes and predict class scores.}
  \label{fig:main_method}
\end{figure*}

In this paper, we present a new multi-modal fusion scheme called \textbf{mmFUSION}. We carefully transform the modalities' features from a higher to a lower size 3D volume by employing separate modality encoders. As shown in parts \textit{A} and \textit{B} of Figure \ref{fig:main_method}, these 3D representations are not bird's eye view (BEV) i.e. most methods compute BEV because of the simplicity which increases the chances of losing geometric structure and semantic density of objects.  We fuse these representations of both modalities through our carefully designed cross-modality followed by a multi-modality attention module. These attention mechanisms use 3D convolution layers followed by a sigmoid for weighting and complementing modality features as depicted in Part \textit{C}. Finally, the decoder layers at the end generate joint multi-modal features (see in part \textit{D}). We attach a 3D head after the decoder of mmFUSION and show the effectiveness of our framework over several early, intermediate, and late fusion methods for the task of 3D object detection.

\section{Related Work}
\label{sec:related_work}
%HERE YOU HAVE TO WRITE AN INTRO ABOUT THE STRUCTURE OF THIS SECTION AND WHAT YOU ARE TALKING ABOUT IN THE SUBSECTIONS.

In this section, we briefly describe 3D detection methods using LiDAR-only, camera-based, and multi-modal fusion schemes which are more relevant to our work.
\subsection{Lidar-based 3D Detection}
Several traditional methods process LiDAR point clouds considering as irregular input and regress 3D boxes in point, voxel, and range view depending on the representation. In the case of point-based 3D detection, the available methods extract, and aggregate features from raw point clouds \cite{yang2019std, shi2019pointrcnn, qi2019deep} before inputting to detection modules. The voxel-based detection methods first form the regular grid from point clouds, and then subsequently use sparse convolution and BEV transformation \cite{graham2017submanifold,chen2022focal,yang2018pixor,yang2018hdnet,lang2019pointpillars,yan2018second,deng2021voxel,yin2021center}. Even some methods project point-clouds to range-view images and process them as images which lack in exploring the geometrical clues because of lack of adjacency while performing transformations \cite{li2016vehicle,fan2021rangedet}. %last sentence may be a conflicting argument because in our case we are also transforming features to predefined space

\subsection{Camera-based 3D Detection}
3D detection from monocular or multi-views is based on image features projected into frustum space directly or with the help of predicted depth \cite{brazil2019m3d,simonelli2019disentangling,wang2021fcos3d,huang2021bevdet,chen2020dsgn}. Since the predicted depth is computationally expensive and also is not an accurate option because of one-to-many mapping and it creates semantic ambiguity, to counter this problem some methods use geometry cues estimated from multi-views \cite{wang2022detr3d, liu2022petr}. In general, alone camera-based 3D detectors are not accurate so far. In our case, we use it as one of the modalities for transforming its extracted multi-level features into a defined 3D space.

\subsection{Multi-modal Fusion for 3D Detection}
As mentioned in the introduction section, the fusion of the multi-sensor can be done at an early, late, and intermediate stage of the 3D detection pipeline. Early fusion is a point-level method of associating modalities features once one of the modalities is transformed into another \cite{ yang2018ipod, ku2018joint, sindagi2019mvx, vora2020pointpainting, liang2018deep, huang2020epnet, yin2021multimodal,li2022voxel}. The late fusion is a way of combining object-level features such as proposals from available modalities \cite{chen2017multi,qi2018frustum, liang2019multi, yoo20203d,huang2020epnet}. On the other hand, if the fusion is happening first at point level and also at the object level, this kind of method is called two-stage detectors \cite{sindagi2019mvx}.

The methods that are quite relevant to our context are \cite{li2022unifying, yoo20203d, chen2022autoalign, sindagi2019mvx, li2022voxel}, where they try to adapt modality fusion based on concatenation or through convolution layers. Whereas our method fuses features of available modalities through carefully designed attention modules minimizing the problem of sensor deficiencies. 
%Meanwhile fusing some specific knowledge between different modalities is another way of modality interaction \cite{hinton2015distilling}.
% while in the discussion, I need to discuss why we didn't compare with these methods because they use extra information such as ROIs or Proposals

%%%%%%%%%%%%%%%%%% adjust these to perfect place
%The significant examples of early fusion schemes are MVX-Net \cite{sindagi2019mvx}, IPOD \cite{yang2018ipod}, Painted PointRCNN \cite{vora2020pointpainting} and AVOD-FPN \cite{ku2018joint}.
%The prominent examples of late fusion are MV3D \cite{chen2017multi}, F-PointNet \cite{qi2018frustum}, UberATG-MMF \cite{liang2019multi} and EPNet \cite{huang2020epnet}.
\section{mmFUSION method}
\label{sec:method}
The mmFUSION framework represented in Fig. \ref{fig:main_method} has four parts: \textit{A}. image features transformation into defined volume limited by detection ranges and voxelization of LiDAR point clouds; \textit{B}. corresponding encoders to transform features into our desired volume space; \textit{C} cross-modality attention and multi-modality attention modules followed by sigmoid to weight modality features; and \textit{D} a decoder to generate desired joint features. 

\subsection{Feature Transformation and Voxelization}
Given the inputs such as an image $I \in R^{W\times H \times 3} $ and LiDAR point clouds $P \in R^{N\times 3}$, we first extract features of both modalities using corresponding backbones. 

For the image, we adapt ResNet50 \cite{he2016deep} followed by feature pyramid network (FPN) \cite{lin2017feature} to extract multi-level features $F_I  \in R^{H \times W \times C}$. Following \cite{rukhovich2022imvoxelnet, philion2020lift}, we transform them into defined space using intrinsic and extrinsic parameters and concatenate them ray-wise to uniformly defined 3D anchors $A \in R^{X^\prime \times Y^\prime \times Z^\prime}$,  where $X^\prime \times Y^\prime \times Z^\prime$ is the volume size. The anchors after associating features $F_I$ can be represented as $A_I \in R^{X^\prime \times Y^\prime \times Z^\prime \times C^\prime}$. Note that we do not adapt depth information as followed by \cite{li2022unifying} which is computationally expensive in this case. 
%rather we follow \cite{rukhovich2022imvoxelnet} but modify it for multi-level features.  

For LiDAR, we voxelize the point clouds adapting dynamic voxelization \cite{zhou2019endtoend} technique to save memory and we employ voxel field encoder (VFE) \cite{zhou2018voxelnet} to create voxel features $V_L \in R^{X \times Y \times Z \times C}$ as shown in the part \textit{A} of Fig. \ref{fig:main_method}.

%Further detail about chosen sparse shapes in order to produce a common low spatial representation is in the proceeding sections.

\subsection{Image and LiDAR Encoders}
To this end, we have transformed image features at defined anchor positions $A_I$ and LiDAR features at voxel coordinates  $V_L$. We employ two separate encoders (Img-Encoder and LiDAR-Encoder) to process corresponding features available in higher volume space, where they adapt their own resolution settings such as for image $S_I \in R^{X^\prime\times Y^\prime\times Z^\prime}$ and LiDAR $S_L \in R^{X\times Y\times Z}$ for sampling the corresponding inputs, and transform these features from higher space volumes to our desired low space volumes of size ${H\times W\times D}$. 
\newline \textbf{Transforming Features into Lower Space Volumes.}
The Img-Encoder is based on stacked 3D convolution layers where the number of layers depends on the ratio between higher to low space volume. It decreases only the depth (z-axis) channel retaining significant information where we follow the similar strategy as used by ImVoxelNet \cite{rukhovich2022imvoxelnet} but we do not produce BEV features. Hence, the Img-Encoder transforms its features from $A_I$ to desired low space volume $V^\prime_I \in R^{H\times W\times D \times C^\prime}$.
The LiDAR-Encoder is based on stacked sparse 3D convolution layers and it generates features of the same size as the image case, which we define as  $V^\prime_L \in R^{H\times W\times D \times C}$.

\textbf{Feature Concatenation.}
In order to fully utilize each modality and learn to complement their deficiencies, we weigh features at each coordinate position in $V^\prime_I$ and $V^\prime_L$. Hence, we adapt concatenating the features of $V^\prime_I$ and $V^\prime_L$, instead of summation or convolution. This process can be expressed as:
\begin{equation}
    V_U = concat(V^\prime_I,V^\prime_L),  
    \label{concate_equation}
\end{equation}
 where $V_U \in R^{H\times W\times D \times (C^\prime + C) }$. 

%This overall process can be summarised as that high-level features are extracted from each modality to have a weighted fusion 

\subsection{Attention Modules and Fusion}
We create cross-modality and multi-modality attention by 
learning the weights for modality information at acquired feature representations. The weighting mechanism is inspired by a gated attention mechanism \cite{fu2019dual, oktay2018attention}.

\textbf{Cross-Modality Attention.}
 Our goal is to take care of complete spaces instead of sampling. We develop a cross-modality attention mechanism tolearn the weighting across the spaces ( $V_I$, $V_L$ ). We carefully design to weigh the modalities features using 3D convolution followed by a sigmoid activation:
\begin{equation}
    V^c_{U} = crossmodAttn(V_U),  
    \label{crossmodAttn_equation}
\end{equation}
where $V^c_{U} \in R^{H\times W\times D \times (C^\prime + C)/2}$ is a weighted cross-modality features.

\textbf{Multi-modality Attention.}
Having weighted features from cross-modality attention, we pass it through a residual block to smooth the embedding. Afterward, we apply another weighting mechanism  to further refine our multi-modal features as: 
\begin{equation}
    V^m_{U} = mmAttn(V^c_{U}),  
    \label{mmAttn_equation}
\end{equation}
 where $V^m_{U} \in R^{H\times W\times D \times (C^\prime + C)/2}$ is a multi-modal feature representation.

 \subsection{Decoder Layers}
The final module in mmFUSION is composed of decoder layer(s) to generate the desired feature representation and is compatible with the 3D detection head. The representation can be expressed as:

\begin{equation}
    mmF = decoder(V^m_{U}),  
    \label{finalconv_equation}
\end{equation}
 where $mmF \in R^{H\times W\times D \times (C^\prime + C)/2}$ are our final joint generated features.

\section{Experiments}
\label{sec:experiments}
We evaluate our proposed mmFUSION on KITTI and NuScenes datasets, where we compare against different fusion strategies as discussed in Sec.  \ref{sec:intro}. We also conduct extensive ablation studies to validate our design choices. 
\subsection{Experimental Setup}
\textbf{Datasets.} We use two widely adopted multi-modality datasets including KITTI \cite{Geiger2012CVPR} and nuScenes\cite{caesar2020nuscenes}. The KITTI dataset contains LiDAR and front-view camera images, where 7,481 samples and 7,518 samples are available for training and testing respectively. The labeled training set is usually split into a train set with 3712 samples and a val set with 3769 samples. The KITTI has an official evaluation protocol for each class considering different levels of difficulties. Such as for Car: IoU = 0.7 is strict, IoU = 0.5 is loose, whereas for both Pedestrian and Cyclist IoU = 0.5 is strict and IoU = 0.25 is loose. We consider average precision (AP) obtained from 40 and 11 recall positions in our evaluation for a fairer comparison against the latest and old methods. 
The NuScenes dataset is collected by six cameras and one top LiDAR, available 700, 150 and 150 scenes for training, validation and testing respectively. Each scene is of 20 seconds in duration and 10 categories are annotated for 3D bounding boxes at 2 Hz in 360 degree field of view. For the NuScenes dataset, we use its official evaluation protocol such as mAP and NDS as evaluation metrics.
\subsubsection{Implementation Details.} 
We trained the mmFUSION framework in an end-to-end manner for 40 epochs using AdamW optimizer with a maximum learning rate of 2$\times$10\textsuperscript{-4}. Our implementation is based on an open-source 3D object detection platform \textbf{(mmdetection3D)} \cite{contributors2020mmdetection3d}. Module-level settings are described in the following sections. Moreover, the training detail is available in the supplementary material. 

\subsubsection{Higher and lower Space Volumes for fusion.}
Our defined higher space volumes for image and LiDAR branches are based on the resolution we set for the features sampling within the detection ranges. For KITTI, we follow the standard detection range which is 0 to 70, -40 to 40, and -3 to 1 on the x, y, and z-axis respectively. Where the size for the image case is $176\times200\times20$ and for LiDAR, we set $1400\times1600\times40$ based on voxel size $0.05\times0.05\times0.1$ meters.
For nuScenes, we set the detection range [-54, 54] for the x-axis and y-axis, and [-5,3] for the z-axis. Similarly, we transform the available 6 multi-view camera features into a volume based on voxel size $0.15\times0.15\times8$ and the LiDAR voxelization based on voxel size $0.075\times0.075\times0.2$ meters. These settings configure the number of layers for both the image and lidar encoders. 

\textbf{Image Encoder.}
Considering KITTI, the image encoder decreases the feature volume size of $176\times200\times20$ to 2 levels ($l$) in the z direction only using a total number of $l$ layers having 3D convolution with stride 2 which is followed by the final $1\times1$ 3D convolution. 
The size of feature volume at the output is $5\times200\times176$ (permuted into LiDAR coordinates). 

\textbf{LiDAR Encoder.} 
LiDAR encoder also follows a similar convention as the image encoder i.e. $l=2$ number of layers but applying the stride = (2, 2, 2) in each direction. Also, note that different from the image encoder, here we use sparse convolution layers. The final output feature volume size is $5\times200\times176$ (already in LiDAR coordinates). 

\textbf{Fusion Modules.}
The mmFUSION fusion block contains one cross-modality attention followed by a multi-modality attention module. The input of the cross-modality module is concatenated feature channels of both modalities such as 128+128 and volume size is the same as $5\times200\times176$ and output mapped to 256 channels. Since the following multi-modality module acts as self-attention, its input 256 channels (output of the previous module) and output channels are also 256.

\textbf{3D Detection Head.}
We attach a simple anchored-based 3D detection head same as used in SECOND \cite{yan2018second}. The 3D head regresses 3D boxes and predicts class scores from \textbf{mmFUSION} features.

%%%%%%%%%%%%%%%%%%%%%%%%%%%%%%%% Experimental Tables %%%%%%%%%%%%%%%%%%%%%%%%%%%%%%%%%
\begin{table}[h!]
    \centering
    \caption{Comparisons of our mmFUSION with different well-known fusion schemes at AP11-IoU=0.7 on KITTI \textit{val} set. (cars Only)}
    \resizebox{\linewidth}{!}
    {\begin{tabular}{lcccc}
    \hline
    \multirow{2}{*}{Method} & \multirow{2}{*}{Fusion} &
    \multicolumn{3}{c}{AP\textsubscript{3D}@Car-R11}   \\
    \cline{3-5}
    &  & Easy & Moderate & Hard\\
    \hline
    MV3D \cite{chen2017multi} & late & 71.30 & 62.70 & 56.60 \\
    F-PointNet \cite{qi2018frustum} & late & 83.76 & 70.92 & 63.65 \\
    MVXNET (VF) \cite{sindagi2019mvx} & 2 stage & 82.30 & 72.20 & 66.80 \\
    MVXNET (PF) \cite{sindagi2019mvx} & 2 stage & 85.50 & 73.30 & 67.40 \\
    AVOD-FPN \cite{ku2018joint} & 2 stage & 83.07 & 71.76 & 65.73 \\
    IPOD \cite{yang2018ipod} & early & 84.10 & 76.40 &75.30\\
    Cont-Fuse (w/o.Geo) \cite{liang2018deep} & feats. & 81.50 & 67.79 & 63.05 \\
    Cont-Fuse \cite{liang2018deep} & feats. & 86.32 & 73.25 & 67.81 \\
    % read  these results in paper pag 10 and defend why these are important to compare because of geometric features used in the method, we don't use such priors
    UberATG-MMF \cite{liang2019multi}& pt-wise (late) & 86.12 & 74.46 & 66.9 \\
    UberATG-MMF \cite{liang2019multi} & roi-wise (late)& 87.93 & 77.87 & 75.58 \\
    UberATG-MMF \cite{liang2019multi} & late & 88.40 & 77.43 & 70.22 \\
    \textbf{mmFUSION-1 (ours) } & feats. & 88.52 & 77.96 & 76.23\\
    \textbf{mmFUSION-2 (ours)} & feats. & \textbf{88.80} & \textbf{78.32} & \textbf{77.21} \\
    \hline
\end{tabular}}
    \label{tab:CarAP11_complete_models}
\end{table}

\begin{table*}[t!]
    \centering
    \caption{Comparisons our mmFUSION with different well-known fusion schemes in different methods at AP40-IoU=0.7 on KITTI \textit{val} set. (cars Only) }
    \resizebox{\linewidth}{!}{
    \begin{tabular}{llcccc cccc}
        \hline 
        \multirow{2}{*}{Method} &\multirow{2}{*}{Fusion Scheme}  &  &
        \multicolumn{3}{c}{\multirow{1}{*}{AP\textsubscript{3D}@Car-R40 (IoU=0.7)}} 
        & & \multicolumn{3}{c}{\multirow{1}{*}{AP\textsubscript{BEV}@Car-R40(IoU=0.7)}} \\
        \cline{4-6}  \cline{8-10}
                        &   & &  Easy&   Moderate&   Hard&    &   Easy&   Moderate&   hard\\
        \hline
        MVXNET (PF) \cite{sindagi2019mvx} &  2 stages  && 84.27 &   72.57 & 68.55 &   &  91.93  &  85.88 & 81.49 \\
        
        Painted PointRCNN \cite{vora2020pointpainting} &  early  & & 88.38 &   77.74 & 76.76 &   &  90.19  &  87.64& 86.71\\
        % we consider 3D-CVF without 3D ROI based refinement to have fair comparison on feature based fusion
        3D-CVF \cite{yoo20203d} &   feats \small{(w/o RoIs)}  &&  89.39 &     79.25 &  78.02 &   & --   &     --&      --\\
        3D-CVF \cite{yoo20203d} &    feats \small{(w RoIs)}   &&  89.67 &     79.88 &  78.47 &   & --   &     --&      --\\
        AutoAlign\cite{chen2022autoalign} &   feats+late   &&  88.16 &     78.01 &  74.90 &   & --   &     --&      --\\
        
        %\hline
        EPNet \cite{huang2020epnet} &     feats \small{(L1 module)}  &     &  89.44 &     78.84 & 76.73 &   & -- &  --    & --  \\

        \textbf{mmFUSION-1 (ours) } & feats. & &  89.57 &   79.80 &   76.95 &   &95.24 &  89.15 & 86.61 \\
        
        \textbf{mmFUSION-2 (ours)} & feats. & &  \textbf{91.04} &   \textbf{80.15} &   \textbf{77.43} &   &   95.01 &  88.77 & 86.36 \\
        
        %EPNet  &            &  92.28&     82.59 & 80.14 &   & 95.51 &      88.76&    88.36\\
        
        %hline
        %\st{PV-RCNN}  &       & 91.53 &     84.36 &  82.29 &  &    92.82 &     90.43&    88.41\\
        %\hline
        %\st{PV-RCNN+VFF}  &       & 92.31&     85.51 &  82.92 & &    95.43&     91.40&    90.66\\
        %\hline
        %\st{Voxel R-CNN}  &       &  92.27&   84.88&   82.50  &  &   95.51&       91.13&   88.85\\
       % \st{Voxel R-CNN + VFF} &  &    89.51&    84.76&   79.21&   &    95.65&       91.75&   91.39\\
        
        %mmFUSION v1 \textbf{(ours)} &  &  87.75 &   75.76&   70.39  &   &    86.47 & 75.28 & 68.05 &   &   94.97 &    87.98 & 83.07 \\
        %mmFUSION v2 \textbf{(ours)} &  &  87.09 &   75.89&   71.14  &   &    85.29 & 75.45 & 68.24 &   &   93.12 &    87.86 & 83.39 \\
        %mmFUSION v3 \textbf{(ours)} &  &  88.44 &   76.61&   73.43  &   &    86.65 & 76.30 & 74.67 &   &   94.17 &    88.33 & 83.93 \\
        \hline
    \end{tabular}
    }
    \label{tab:car_results_AP40_latest}
\end{table*}

\begin{table*}[t!]
\centering
\caption{Comparison of mmFUSION with recent methods of early, late, and intermediate (feats.) level fusion on nuscenes \textit{val set}. We compare overall and at per class level as well.}
\resizebox{\linewidth}{!}
{\begin{tabular}{llcccccccccccc}
\hline
Method & Fusion Schemes &\textbf{NDS(\%} & \textbf{mAP(\%)} & Car & Truck&C.V& Bus& Trailer & Barrier &Motor.& Bike&Ped.& T.C.\\
\hline
FusionPainting\cite{xu2021fusionpainting}&early&67.41&62.15&83.98&59.97&22.90&71.13&41.53&63.29&66.85&54.35&82.86&74.66\\
FUTR3D\cite{chen2023futr3d}& feats. + late &68.30&64.50&86.30&61.50&26.00&71.90&42.10&64.40&73.60&63.30&82.60&70.10\\
VFF\cite{li2022voxel}&feats.&68.70&63.40&86.40&60.10&23.80&71.70&38.80&68.00&71.50&52.20&86.20&75.80\\
mmFUSION \textbf{(ours)} &feats.&\textbf{69.75}&\textbf{65.43}&\textbf{87.60}&60.79&\textbf{28.50}&\textbf{72.62}&40.62&67.24&\textbf{74.02}&\textbf{65.17}&85.24&72.46\\
Improvement $\uparrow$ & &1.43\% &3.10\%&1.36\%&1.13\%&16.49\%&1.26\%&4.48\%&-1.13\%&3.40\%&19.90\%&-1.12\%&-4.60\%\\
\hline
\end{tabular}}
\label{tab:NuScenes_val_on_class_level}
\end{table*}

\begin{table*}[t!]
    \centering
    \caption{Comparison of mmFUSION with UVTR \cite{li2022unifying} on nuscenes \textit{val set}}
    \resizebox{\linewidth}{!}{
    \begin{tabular}{llccccccc}
         \hline
         Method & Fusion Scheme& \textbf{NDS(\%} & \textbf{mAP(\%)} & mATE $\downarrow$& mASE $\downarrow$& mAOE $\downarrow$ & mAVE $\downarrow$ & mAAE $\downarrow$ \\
        \hline
         UVTR\cite{li2022unifying}&feats.&70.20&65.40&0.333&0.258&0.270&0.216&0.176\\
         mmFUSION \textbf{(ours)} &feats.&69.75&\textbf{65.43}&\textbf{0.327}&\textbf{0.256}&\textbf{0.268}&0.219&\textbf{0.170}\\
         Improvement &&--&--&1.83\%&0.78\%&0.74\%&-1.37\%&3.52\%\\
         %Improvement &&-0.64\%&0.04\%&1.83\%&0.78\%&0.74\%&-1.37\%&3.52\%\\
         \hline
    \end{tabular}
    }
    
    \label{tab:NuScenes_val_errors}
\end{table*}

\begin{table*}[h]
\centering
\caption{Ablation experiments on KITTI validation set (cars Only) where SLim: single level image features, MLim: multi-level image features, SC: simple concatenation, JFG: joint feature generation, mmFD: increasing depth (adding mmFUSION modules at more defined spaces), mmFL: increasing mmFUSION length.}
\resizebox{\linewidth}{!}
{\begin{tabular}{lcccccc cccc cccc cr}
\hline
\multirow{2}{*}{Method}
&\multirow{2}{*}{SLim}
& \multirow{2}{*}{MLim}
& \multirow{2}{*}{SC}
& \multirow{2}{*}{JFG}
& \multirow{2}{*}{mmFD}
& \multirow{2}{*}{mmFL} 
&
&\multicolumn{3}{c}{AP\textsubscript{3D}@Car-R40 (IoU=0.7)}
& 
&\multicolumn{3}{c}{AP\textsubscript{BEV}@Car-R40(IoU=0.7)}
&
&\multirow{2}{*}{$\Delta$} \\
\cline{9-11}  \cline{13-15}
& & & & & & & & Easy&   Moderate&   hard & & Easy&   Moderate&   hard &&\\
\hline
%Baseline  & & & & & & & & 93.46& 90.10&87.47 & & 84.27 &   72.57 & 68.55 &   &  91.93  &  85.88 & 81.49 \\
Baseline* & &\checkmark& & & & & & 82.86 &   69.82 & 65.47 &   & 92.07   &  85.86 & 81.58 &&-\\
\hline
Config-A& \checkmark & & \checkmark& & & & &  87.40 &   75.10 &   70.39  &   &   94.97 &   85.88  & 81.00 && +7.56\% \\
Config-B & \checkmark & & &\checkmark & & & &  87.09 &   75.09&   71.14  &   &  93.12 &    87.86 & 83.22 && +7.54\%\\
Config-C &  & \checkmark & &\checkmark & &  &  & 88.44 &   76.61&   73.43   &    &   92.96 &    86.22 & 83.66 && +9.72\% \\
%v3 (ours) &  &\checkmark & &\checkmark & & & & & 96.18 & 90.87 & 88.17 & & 88.44 &   76.61&   71.94   &    &   94.17 &    88.33 & 83.93  \\
Config-D &  &\checkmark & &\checkmark &1 & &  & 89.57 & 79.80 &  76.95  &    &   94.88 &    88.73 & 86.11 && +14.29\% \\
Config-E &  &\checkmark & &\checkmark & &\checkmark &  & 88.88 & 79.71 &   76.94  &    &   93.51 &    88.82 & 86.32 && +14.16\% \\
Config-F (final) &  &\checkmark & &\checkmark & 2 & & & \textbf{91.04} & \textbf{80.15} &  \textbf{77.43} &    &   \textbf{95.01} &  \textbf{88.77} & \textbf{86.36} && \textbf{+14.79}\% \\
\hline
\end{tabular}}
\label{tab:ablation_study}
\end{table*}

\begin{table*} [h]
    \centering
    \caption{Latency and performance of different mmFUSION configurations on KITTI \textit{val set}.}
    \resizebox{\linewidth}{!}
    {\begin{tabular}{lccccc cccc cccc cr}
        \hline 
        \multirow{2}{*}{Configs.}&  
        \multirow{1}{*}{Feats. Shape}&
        \multirow{1}{*}{Img Voxel Size}&
        \multirow{1}{*}{Pts Voxel Size}&
        \multirow{1}{*}{mmFD}&
        \multirow{1}{*}{Latency}&&
        \multicolumn{3}{c}{\multirow{1}{*}{AP\textsubscript{3D}@Car-R40 (IoU=0.7)}} &  
        & \multicolumn{3}{c}{\multirow{1}{*}{AP\textsubscript{BEV}@Car-R40 (IoU=0.7)}}  &&
        \multirow{3}{*}{$\Delta$}  \\
        \cline{8-10}  \cline{12-14}
        & $H \times W \times D $& (m) & (m) & & (ms)&  &   Easy&   Moderate&   hard & & Easy&   Moderate&   hard &&\\
        \hline
        \multirow{1}{*}{Tiny} &
        \multirow{1}{*}{42$\times$42$\times$5} &
        \multirow{1}{*}{1.67$\times$1.67$\times$0.25} & 
        \multirow{1}{*}{0.20$\times$0.20$\times$0.10}&
        %1 & xx &  &  &  & &   &    &     &   &    &     &   \\
        %&  &  & &
        2 & 5 &  &  55.72 &   45.63 &   43.07  &   &   63.94 &   55.69  & 54.22 &&-43.07\%  \\

        \multirow{1}{*}{Small} &
        \multirow{1}{*}{86$\times$86$\times$5} &
        \multirow{1}{*}{0.81$\times$0.81$\times$0.25 }&
        \multirow{1}{*}{0.10$\times$0.10$\times$0.10} & 
        %1 & xx &  &  &  &&   &   &    &   &   &  &  \\
        %&  &  & & 
        2 & 20 & &  85.48 &  72.93 &  68.76  &   &  91.95 & 83.38 & 80.76 &&-9.01\% \\
        
        \multirow{1}{*}{Base} &
        \multirow{1}{*}{174$\times$174$\times$5} &
        \multirow{1}{*}{0.40$\times$0.40$\times$0.25}&
        \multirow{1}{*}{0.05$\times$0.05$\times$0.10}& 
        %1 & xx &  & -- & -- &-- &  &  -- &  -- & --   &    & +XX\%    \\
        %&  & & &
        2  & 33 & & 89.57 &   77.48&   74.55   &    &   93.46 &    86.64 & 83.91  && -3.33\% \\
        
        \hline
        \multirow{3}{*}{Base (final)}&
        \multirow{3}{*}{175$\times$200$\times$5}  & 
        \multirow{3}{*}{0.40$\times$0.40$\times$0.25} &
        \multirow{3}{*}{0.05$\times$0.05$\times$0.10} & 
        1 &  53 & & 89.57 & 79.80 &  76.95 &    &   95.24 &   89.15 & 86.61 &&--0.436\%  \\
        &  & &  & 
        2 in len &  65 && 89.16 &  79.71 & 76.94  &&
        93.02 & 88.72 & 86.31&&-0.548\%  \\
        &  & &  & 
        2 &  58 && 91.04 & 80.15 & 77.43 & &
        95.01 & 88.77 & 86.36 &&100\%  \\
        \hline
    \end{tabular}}
    \label{tab:ablation_study_mmfusion_scale}
\end{table*}

\begin{table*}[h]
    \centering
    \caption{Comparisons of our mmFUSION framework with baseline \cite{sindagi2019mvx} and prior different fusion scheme on KITTI \textit{test} set.}
    \resizebox{\linewidth}{!}
    {
    \begin{tabular}{llccccccc}
        \hline
        \multirow{2}{*}{Method} &   \multirow{2}{*}{Fusion Scheme} &
        \multicolumn{3}{c}{\multirow{1}{*}{AP\textsubscript{3D}@Car (IoU=0.7)}} & 
        &  \multicolumn{3}{c}{\multirow{1}{*}{AP\textsubscript{BEV}@Car (IoU=0.7)}}\\
        \cline{3-5} \cline{7-9}
        & & Easy & Moderate & Hard & & Easy & Moderate & Hard  \\
        \hline
        MV3D \cite{chen2017multi}& Late & 74.97 & 63.63 & 54.00 & & 86.62 & 78.93 & 69.80 \\
        IPOD \cite{yang2018ipod} & Early & -- & -- & -- & & 89.64 & 84.62 & 79.96 \\
        Cont-Fuse \cite{liang2018deep} &  Feats. & 82.50 & 66.20 & 64.04 & & 88.81 & 85.83 & 77.33  \\ % check again for its 40 point verification
        F-PointNet \cite{qi2018frustum} & Late & 82.19 & 69.79 & 60.59 & & 91.17 & 84.67 & 74.77  \\
        AVOD-FPN \cite{ku2018joint} & 2 Stages & 83.07 & 71.76 & 65.73 & & 90.99 & 84.82 & 79.62 \\
        MVXNet (PF) \cite{sindagi2019mvx} & 2 Stage & 83.20 & 72.70 & 65.20 & & 89.20 & 85.90 & 78.10 \\ 
        %F-ConvNet & Early & \textbf{87.36} & \textbf{76.39} & 66.69 && 91.51& 85.84 & 76.11 \\
        % we  do not consider 3D CVF in test comparison because of  its 3D RoI based refinement mechanism which do not make fair  with us
        %UberATG-MMF & mid & 88.40 & 77.43 & 70.22 & & -- & -- & --  \\
        
        %3D-CVF & Feature & 89.20 & 80.05 & 73.11 & & -- & -- & -- \\
        %EPNet & Feature & 89.81 & 79.28 & 74.59 & & 94.22 & 88.47 & 83.69\\
        \textbf{mmFUSION (ours)} & Feats. & \textbf{85.24} & \textbf{74.38} & \textbf{69.43} & & \textbf{90.35} & 84.60 & \textbf{79.82} \\
        \hline
    \end{tabular}
    }
    \label{tab:kitti_test_results}
\end{table*}

\subsection{Main Results}
\textbf{KITTI Results.}
In Table \ref{tab:CarAP11_complete_models}, we report mmFUSION performance against the methods having AP11 protocol for 3D detection on the validation set. We use the MVXNet \cite{sindagi2019mvx}, a two-stage detector (an early and late fusion strategy) as the first baseline, and several other fusion schemes methods for comparison. The mmFUSION outperforms point and voxel-based fusion settings of MVXNet, moreover, it surpasses all other fusion schemes such as MV3D, F-PointNet, UberATG-MMF, and IPOD. Further, mmFUSION performs better than feature-based schemes such as Cont-Fuse considering its different settings (with and without) its geometric and KNN features, and UberATG-MMF with its different settings (point or ROI-wise fusion scheme).

We also evaluate mmFUSION performance over methods having AP40 criteria (a newly adapted metric by KITTI and followed by lateral methods). In Table \ref{tab:car_results_AP40_latest}, we report a comparison with a strong early fusion method such as Painted PointRCNN \cite{vora2020pointpainting}, a feature-based fusion of 3D-CVF \cite{yoo20203d}, and L1 module for feature fusion of EPNet \cite{huang2020epnet}. Since 3D-CVF uses a 3D RoI-based refinement mechanism before final proposals which does not make it a fair comparison with mmFUSION, but still mmFUSION outperforms all its settings and from EPNet as well. Overall we observed that several different fusion schemes use 2D or 3D ROIs/proposals to help multimodal fusion, instead, mmFUSION does not use ROIs, depth, or even pre-trained on single modalities. We train mmFUSION end-to-end which is a pure features-based fusion scheme on image and LiDAR features and attaching a simple 3D head surpasses all prior fusion schemes. 
We also report results on the kitti test set for car benchmark in Table \ref{tab:kitti_test_results}, mmFUSION performs better than previous fusion schemes and the baseline \cite{sindagi2019mvx} by 2.25\% \\

\textbf{NuScenes Results.} 
In Table \ref{tab:NuScenes_val_on_class_level}, we compare well-known early, late, and intermediate level (feature level) fusion methods with mmFUSION on the nuscenes dataset, where mmFUSION surpasses all these approaches. Performance gain brought by mmFUSION over the latest feature-based scheme such VFF\cite{li2022voxel} is 1.43\% NDS and 3.10\% mAP. Further in Table \ref{tab:NuScenes_val_errors}, We report mmFUSION results of 3D detection errors on nuscenes dataset, and we also compare with UVTR \cite{li2022unifying}, it is a feature-based fusion method similar to VFF \cite{li2022voxel}, we observe mmFUSION improvement with 1.83\% in mATE, 0.78\% in mASE, 0.74\% in mAOE, and 3.52\% in mAAE.  

More details about the results are present in the supplementary material. 
%\textcolor{red}{Temporary Results} \\
%We implemented the baseline comparison method \cite{sindagi2019mvx}, and the base components of our proposed pipeline without the aggregator. Our first  test experiment is based on Kitti dataset \cite{Geiger2012CVPR} which contains three difficulty levels(easy, moderate, and hard) in the annotations. Our results show a mean average precision (mAP) improvement for two categories: cars from 85.77 to 86.63 and pedestrians from 60.55 to 63.14. However, these are only temporary results, and the key component of our proposed pipeline, the aggregator, is still missing and should in principle lead to better performances.

\subsection{Ablation Study}
We conduct extensive ablation studies on the KITTI validation set to deeply analyze and evaluate the effectiveness of each module in \textbf{mmFUSION}. Moreover, we analyze the model scale by conducting experiments on different configurations. To put a reference throughout the study, we use an anchored 3D head as a box regressor and class score predictor. We provide details about each experiment with selected modules in the proceeding sections and analysis report in Table \ref{tab:ablation_study}. The $\Delta$ shows the performance gain (in moderate difficulty level) while adding a specific module into the pipeline. The performance gain of \textbf{mmFUISON} with its all modules compared to the baseline is \textbf{14.79\%}. \\

\textbf{Baseline*.} Considering a two-stage detector \cite{sindagi2019mvx} as the baseline, we remove its point or voxel-based fusion (early fusion) and sum the modality features before the detection head. This can be referred to as naive feature base fusion and we refer to it as Baseline*. \\
 \textbf{Config-A: Simple Concatenation (SC).}
We start our ablation study by simply concatenating features (SCF) extracted from each modality's encoders and there are no cross or multi-modal attention modules in this choice. We use the single-level image (SLim) features from the image backbone as input to the image encoder. \\
\textbf{Config-B: Joint Feature Generation.}
To test the compatibility with the 3D head, we added a joint feature generation layer after adapting config-A. This configuration does not boost the performance but makes the framework compatible with a desired 3D head.\\
\textbf{Config-C: Using Multi-level Image Features.}
In order to differentiate whether multi-level image (MLim) features have performance benefits over SLim in our proposed framework, we replace SLim with MLim in this experiment. We observe 9.72\% gain in performance. \\
\textbf{Config-D: Adding mmFUSION Module.}
Keeping config-C and adding the mmFUSION module at our defined space ($5\times200\times176$) boosts sufficient performance such as 14.29\%. This shows the effect of cross-attention and multi-modal attention mechanism choice in our mmFUSION framework.  \\
\textbf{Config-E: Increasing mmFUSION Length}
In this choice, we ablate that adding attention modules sequentially does not increase performance as it is almost the same as configuration D. Hence, we keep a single cross-modality and single multimodality module in one mmFUSION module for future choices. \\
\textbf{Config-F: Increasing mmFUSION Depth}
Since we can acquire features at multiple levels from the image and LiDAR encoders, we ablate to add two mmFUSION modules one at ($5\times200\times176$) and another at ($10\times200\times176$), and observed an increment in performance. We also observe that getting more than two spaces for fusion is computationally too expensive. On the other hand, increasing length does not boost performance we we see in the last configuration choice. Hence configuration F is the optimal choice of the mmFUSION framework.

%\textbf{Config-D: Adding Attention Mechanism.}
%
%We check the effects of self-attention composed of linear layers on generated joint multi-modal features. For this purpose, we reshape the features before and after the attention layer. We observe increasing layers considering our 3D spatial shape is not feasible computationally. 

%\textbf{Config-E: Cross-Modality Attention and Multi-Modal Feature Generation.}
%
%Since the attention mechanism is computationally an expensive task, The features with 3D spatial shapes are very hard to process with cross-modality as we need to consider both modalities simultaneously. In order to check cross-attention (CA) effects, we use criss-cross attention layers \cite{huang2019ccnet} (a computationally cheap option) on both modalities as depth-wise (in z-axis) in this experiment. 

%\textbf{Config-F: Feature Generation with mmFUSION Module.}
%
%This experiment is our best-proposed module which contains cross-modality attention followed by another attention on acquired multi-modality features as explained in the main methodology of our paper. 

%[FIXME: there is no discussion of these ablations studies! How these have drive the model design? Why and what works better? Which is the reason? etc...]
\textbf{Model Scale and Inference Latency.}
We compare the performance of different configurations of our mmFUSION in Table \ref{tab:ablation_study_mmfusion_scale}. We ablate the different scales of mmFUSION in terms of the defined volume of feature maps to verify the performance accordingly and compare inference latency. We noticed the tiny and small versions having lower volume size of $42\times42\times5$ and $86\times86\times5$ reduces computational and memory burden but same time reduces 43\% mAPs and 9\% mAPs as well (compared moderate difficulty level). The base version with a feature volume size $174\times174\times5$ is close to the final version $176\times200\times5$ but a little less in performance. Hence it is a trade-off between the chosen configuration and the required performance. Moreover, we also observe increasing mmFUSION modules in a sequential manner does not create a significant difference, but adding modules at different levels of encoder features increases some performance.

\subsection{Qualitative Evaluation}
\begin{figure*}[h!]
    \centering
    \begin{subfigure}[b]{0.44\textwidth}
            \centering
            % raw image
            \begin{subfigure}[b]{1\textwidth}
                \centering
                \includegraphics[width=\textwidth]{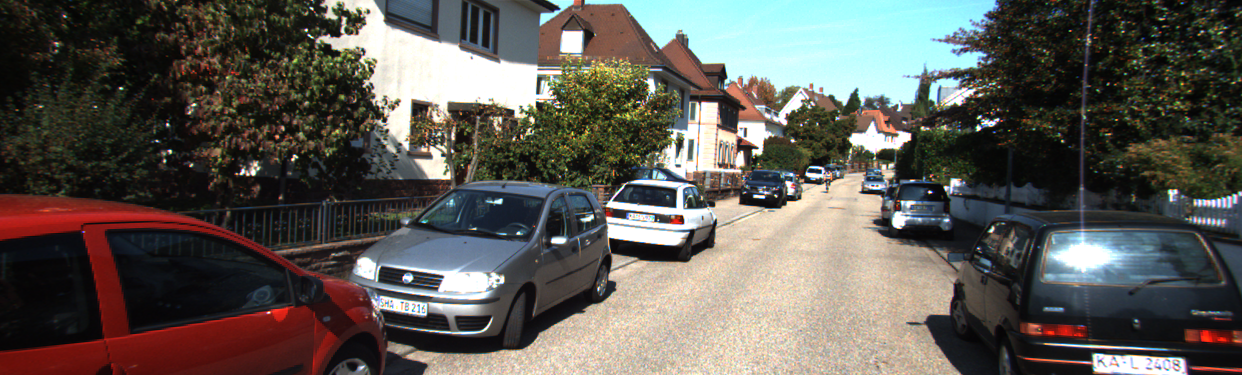}
            \end{subfigure}
            % lidar points
            \begin{subfigure}[b]{1\textwidth}
                \centering
                \includegraphics[width=\textwidth]{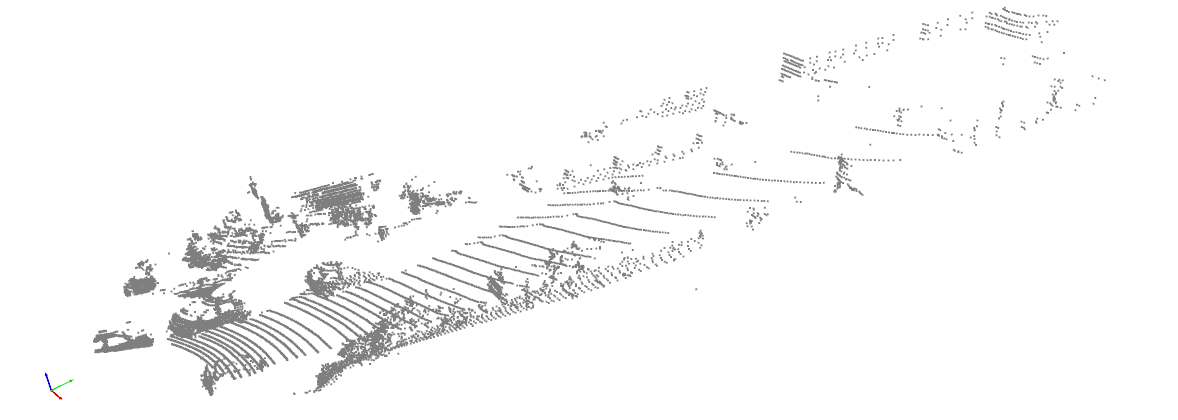}
            \end{subfigure}
            % projected points on image
            \begin{subfigure}[b]{1\textwidth}
                \centering
                \includegraphics[width=\textwidth]{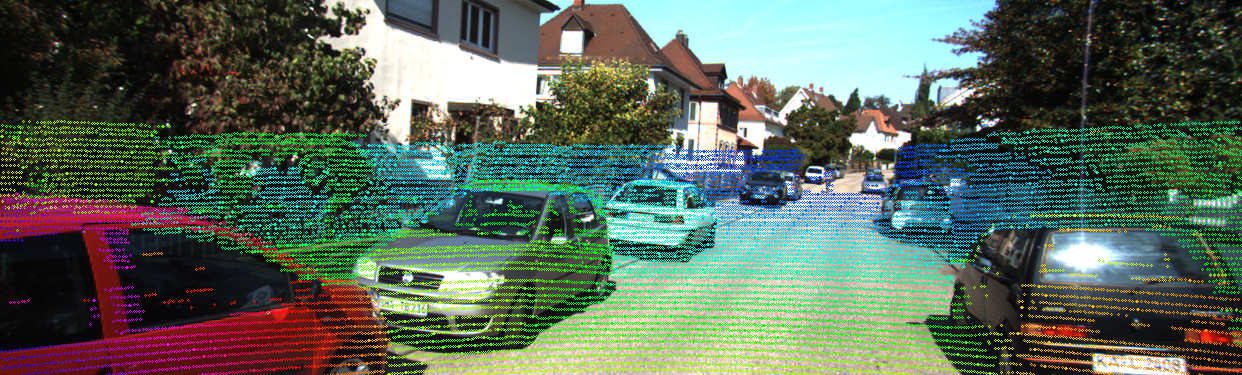}
             %\vspace{0.3cm}
            \end{subfigure}
           
            % projected gt 3D boxes on image
            \begin{subfigure}[b]{1\textwidth}
                \centering
                \includegraphics[width=\textwidth]{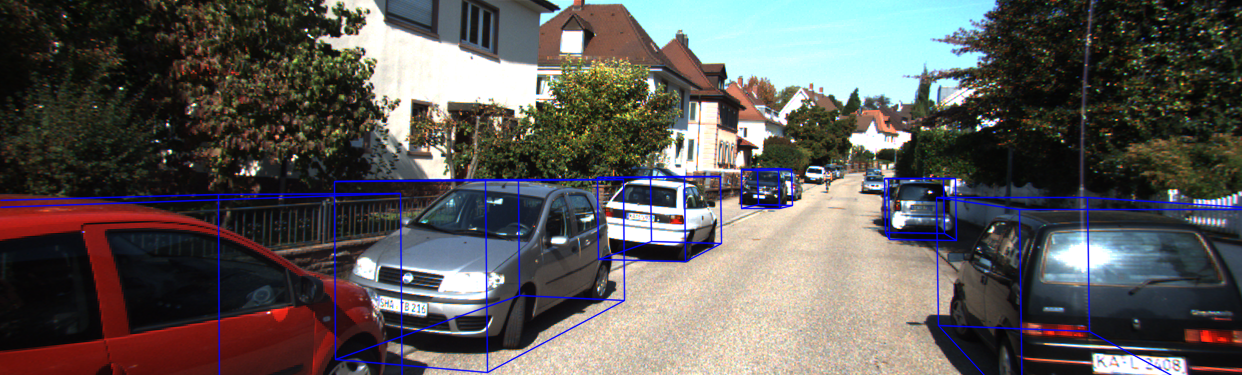}
            \end{subfigure}
            % projected predicted 3D boxes on image
            \begin{subfigure}[b]{1\textwidth}
                \centering
                \includegraphics[width=\textwidth]{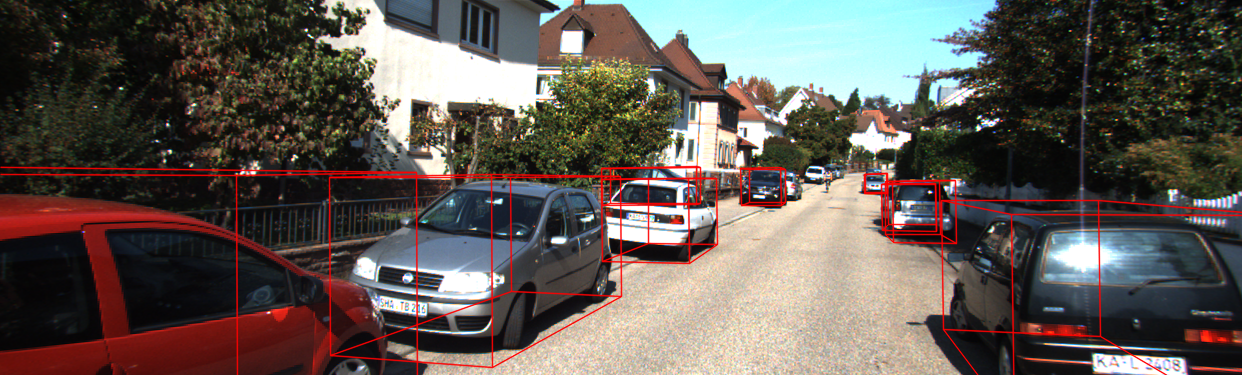}
            \end{subfigure}

    \end{subfigure}
    %\hspace{-0.1cm}
    \begin{subfigure}[b]{0.41\textwidth}
             \centering
            % img low spatial shape
            \begin{subfigure}[b]{0.95\textwidth}
                \centering
                \includegraphics[width=\textwidth]{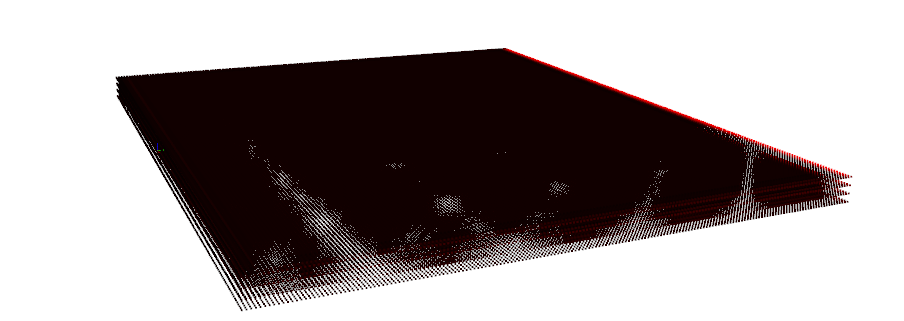}
            \end{subfigure}
            % lidar low spatial shape
            \begin{subfigure}[b]{0.95\textwidth}
                \centering
                \includegraphics[width=\textwidth]{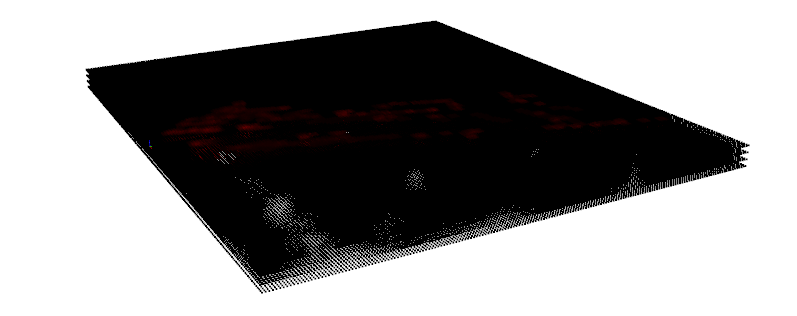}
            \end{subfigure}
            % multimodality low spatial shape
            \begin{subfigure}[b]{1\textwidth}
                \centering
                \includegraphics[width=\textwidth]{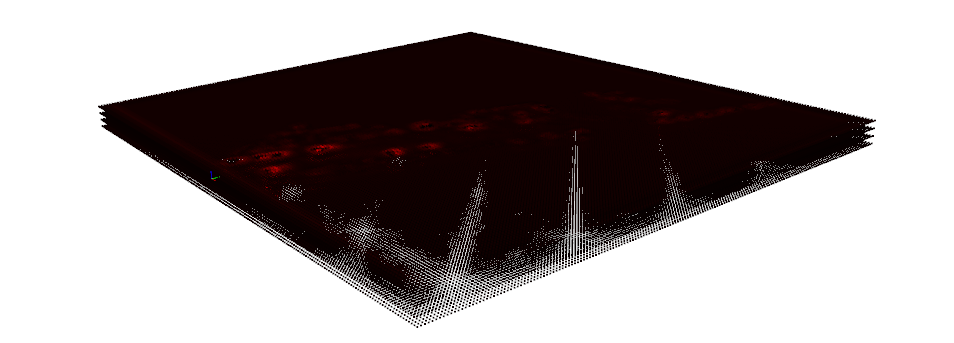}
                %\vspace{0.3cm}
            \end{subfigure}
            % drawn gt boxes on multimodality low spatial shape
            \begin{subfigure}[b]{1\textwidth}
                \centering
                \includegraphics[width=\textwidth]{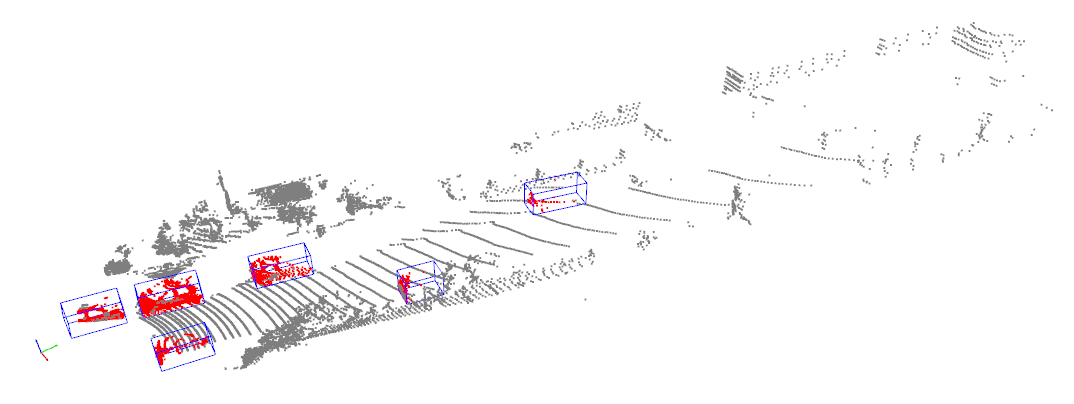}
            \end{subfigure}
            % drawn pred boxes on multimodality low spatial shape
            \begin{subfigure}[b]{1\textwidth}
                \centering
                \includegraphics[width=\textwidth]{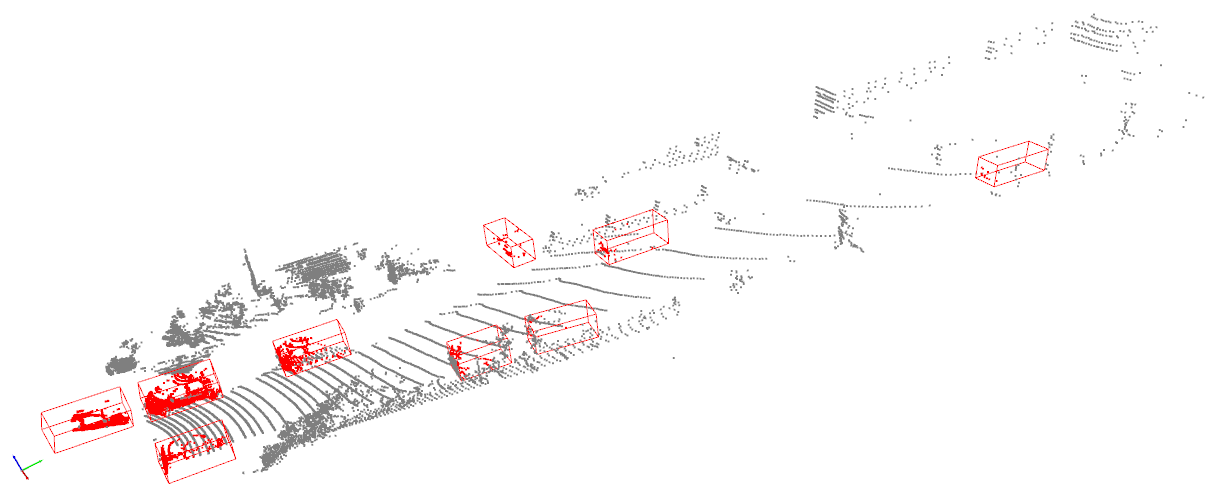}
            \end{subfigure}
            
    \end{subfigure}
    \caption{Visualization of mmFUSION on KITTI example. First row both columns: image, and its features in low space volume. Second: LiDAR point clouds, and their features in the low space volume. Third: projected LiDAR points on the image, and the figure to its right side showing mmFUSION features after cross-modality and multi-modality attention block (the color shows L1 norm value, where the black is almost 0, and the extreme red shows the highest norm value). Fourth and Fifth: ground truth 3D boxes \color{blue}{(in blue)} \color{black} and predicted 3D boxes \color{red}{(in red)}.}
%    \caption{Examples of the proposed low spatial shape on KITTI \textit{val} set. The blue points indicate the projected features in low spatial shape while green points show LiDAR points.}
    \label{fig:overll_qualitative_example1}
\end{figure*}

This section shows qualitatively how the proposed mmFUSION framework detects objects leveraging complementary information from both sensors. In Figure \ref{fig:overll_qualitative_example1}, we first visualize image and its corresponding features in lower sized volume (output of proposed image encoder). Then we show lidar points and their corresponding features in the lower-sized volume (output of proposed LiDAR Encoder). Note that red regions show the L1 norm of the feature vector at that particular location. It can be seen in the case of images, features are available but spread, whereas in the case of LiDAR, they seem more localized showing the presence of objects (the red regions). Third, we show the image with projected LiDAR points (just to show both modalities together), and towards its right side, the final joint features in lower feature volume (output of mmFUSION modules: cross and multi-modality attentions). It can be seen in this representation how objects' regions (red regions) are significantly localized. Hence, we see how objects that are either far or not even having LiDAR points are still detected correctly because mmFUSION leverages information from the camera for those points and fills the gap between both modalities. It is also important to notice visually some objects are not annotated in the KITTI dataset but are still correctly detected by the mmFUSION framework.
%We show an example of detected 3D objects with, network and defined spaces details in figure \ref{fig:overll_qualitative_example1}.

\section{Conclusions and Discussion}
\label{sec:conclusions}
We presented a new method for multi-modality fusion called \textbf{mmFUSION} for 3D object detection. Our key innovation lies in that we propose a feature-level fusion that does not rely on either ROIs or 2D/3D proposals from modalities. The image and  LiDAR features from higher-sized volumes are transformed into lower-volume by image and LiDAR encoders respectively. The following multi-modal fusion module is based on cross-modality and multi-modality attention mechanisms which learn to weigh the modalities. The decoder layers at the last stage generate the desired size of multi-modal features for a 3D head. 
The first key advantage of the mmFUSION framework is to avoid lower-level feature interaction (early fusion problems) i.e. there is a lack of adjacency between transformed image features and available 3D points. The second advantage is to introduce an independent fusion scheme such as avoiding deficient proposals (late fusion problems) from any modality. Additionally, we have a one-stage fusion strategy that is very adaptable for multi-modal 3D detectors. 
The mmFUSION can also be trained on both modalities simultaneously instead of having pre-trained models on single modalities. The mmFUSION is a more adaptive fusion scheme, which can serve as a new feature-based fusion baseline for multi-modal 3D detection in the future. 
The limitation of mmFUSION is, not to acquire a higher size of feature volumes from image and LiDAR encoders. Giving large feature volumes to cross-modality and multi-modality attention modules can create a computational burden while computing attention weights.

{
    \small
    \bibliographystyle{ieee_fullname}
    \bibliography{main}
}

\end{document}